\documentclass{article}

\usepackage{arxiv}

\usepackage[utf8]{inputenc} 
\usepackage[T1]{fontenc}    
\usepackage{hyperref}       
\usepackage{url}            
\usepackage{booktabs}       
\usepackage{nicefrac}       
\usepackage{microtype}      
\usepackage{lipsum}
\usepackage{graphicx}
\graphicspath{./images/}
\usepackage{float}
\usepackage{subcaption}

\begin{document}
\title{Finding the Ground-Truth from Multiple Labellers: Why Parameters of the Task Matter}

\author{
  Robert McCluskey\\
  Data Science Department, Caspian \\
  Newcastle-upon-Tyne, UK \\
  \texttt{ROB.MCCLUSKEY@CASPIAN.CO.UK} \\
   \And
 Amir Enshaei \\
  Data Science Department, Caspian \\
  Wolfson Childhood Cancer Research Centre \\
  Newcastle-upon-Tyne, UK \\
  \texttt{AMIR.ENSHAEI@NEWCASTLE.AC.UK} \\
  \AND
  Bashar Awwad Shiekh Hasan\\
  Data Science Department, Caspian \\
  Newcastle-upon-Tyne, UK \\
  \texttt{BASHAR.HASAN@CASPIAN.CO.UK} \\
}

\maketitle
\begin{abstract}
Employing multiple workers to label data for machine learning models has become increasingly important in recent years with greater demand to collect huge volumes of labelled data to train complex models while mitigating the risk of incorrect and noisy labelling. Whether it is large scale data gathering on popular crowd-sourcing platforms or smaller sets of workers in high-expertise labelling exercises, there are various methods recommended to gather a consensus from employed workers and establish ground-truth labels. However, there is very little research on how the various parameters of a labelling task can impact said methods. These parameters include the number of workers, worker expertise, number of labels in a taxonomy and sample size. In this paper, Majority Vote, CrowdTruth and Binomial Expectation Maximisation are investigated against the permutations of these parameters in order to provide better understanding of the parameter settings to give an advantage in ground-truth inference. Findings show that both Expectation Maximisation and CrowdTruth are only likely to give an advantage over majority vote under certain parameter conditions, while there are many cases where the methods can be shown to have no major impact. Guidance is given as to what parameters methods work best under, while the experimental framework provides a way of testing other established methods and also testing new methods that can attempt to provide advantageous performance where the methods in this paper did not. A greater level of understanding regarding optimal crowd-sourcing parameters is also achieved.
\end{abstract}

\keywords{
    Labels
    \and
    Crowdsourcing
    \and
    Expectation-maximisation
    \and
    CrowdTruth
    \and
    Majority vote
}

\section{Introduction}

\par
For supervised learning, the gathering of precisely labelled data is crucial to the success of any model, although it is often the most labour-intensive part of any project that requires external resources\cite{Roh2018ASO}. For highly specialised tasks, Subject Matter Experts (SME) are often employed to provide labels, while crowd-sourcing platforms such as Mechanical Turk (MTurk) have also become popular for gathering high volumes of data quickly, when the task can be completed by a layman, as can be seen in the growth of models like ImageNet\cite{Russakovsky2015ImageNetLS} and their ability to sometimes provide labels similar to that of a “gold standard”\cite{snow-etal-2008-cheap}. Obtaining this “gold standard” is crucial to the production of a high-performance model.
\par
As well as increasing the speed at which data is gathered, having multiple workers label the same data can remove the problem of a single worker being considered infallible, thus reducing the risk of incorrect labels due to human error, whether it be due to lack of expertise in labelling a particular data point, misinterpreting a task\cite{Le10ensuringquality} or applying the incorrect label accidentally. There also remains the possibility that workers simply have different interpretations of the labels they apply\cite{Giancola2018PermutationInvariantCO}. Whether it is crowd-sourced workers or experts, we still run into an issue of which workers we should trust in what instances, and inferring which label can be considered the "ground-truth" for each datum in sample becomes a problem when workers give different labels to a particular data point. While there are methodologies to resolve these problems, little is known about the circumstances under which these methods provide the most benefit and whether some methods are better than others under certain task conditions.\par

Crowd-sourcing platforms have grown in popularity due to the ease of gathering data for a relatively low cost. However, accuracy from each worker is not assured and the anonymity of the workers means that the person who wants the data must put blind faith into others who may provide poor quality labels. Halpin \& Blanco\cite{Halpin2012MachineLearningFS} considered two different types of poor quality workers: those who deliberately select answers as quickly as possible in order to get paid - referred to as "bad faith" workers - and those who are "unsuitable" workers, due to their lack of understanding or poor ability. The "bad faith" workers may also be known as "spammers"\cite{Raykar2012EliminatingSA} or “adversarial” workers\cite{Jagabathula2017IdentifyingUA}. The growth of crowd-sourcing platforms has come at the cost of more malicious workers sabotaging machine learning projects for financial gain\cite{Wang2014ManVM}. Detecting these “bad faith” workers is of critical concern for the users of crowd-sourcing platforms, as they waste time and money while also negatively impacting the final decision. On the other hand, some workers may have misunderstood the task or have limited experience of labelling the given data, which can be provided with stronger guidance or training. However, when working with a high number of anonymous workers, it is difficult to disseminate between the two. Detecting poor workers early in a data gathering exercise is of the utmost importance when considering efficiency and cost effectiveness\cite{Lee2018EffectiveQA}.
\par
Crowd-sourcing platforms, where a high number of workers can be employed, are most suitable for tasks that can be completed by a layman, but tasks requiring a high level of expertise often need SMEs. Owing to a smaller pool of highly skilled expertise, recruiting SMEs is more challenging and costly\cite{Aroyo2013CrowdTH} and crowd-sourcing platforms may not be able to provide access. When a task is of a certain degree of difficulty or complexity, experts can often disagree among themselves - something that is often seen in the diagnosis of medical conditions\cite{Benbadis2010TheTO,Graber2013TheIO,Ehrich2018TheIO} due to the highly specialised nature of the task, and reliance on individual professional experience, adding more difficulty to the data gathering process. As an example, compare a task where a worker must label whether an image is a cat or a dog with a task where a worker must label whether a cancer diagnosis is benign or malignant. The former task has more people who will likely be able to contribute knowledgeably, while the latter necessitates specialised training and expertise. Even in these tasks where we consider people as “experts”, seeing noise in label sets is inevitable and this can have a severe impact on any model if not controlled correctly\cite{Karimi2020DeepLW}. One possible control is employing a single expert, or an “oracle”, to make final decisions on data entries where there is disagreement amongst non-experts\cite{Dolatshah2018CleaningCL}. This, however, creates an extra overhead and also relies upon the knowledge of the “oracle” to be entirely accurate, which is a flawed concept and does not entirely rule out errors in the final label set.
\par
One of the reasons for disagreements in expert labelling is that even within different fields of expertise, some are better at identifying certain things than others. Additionally, inherent subjectivity ensures there will always be some variation in answers\cite{Raykar2010LearningFC}, as well as bias within each worker\cite{Hube2018LimitBiasMW}. Valizadegan, Nguyen \& Hauskrecht\cite{Valizadegan2013LearningCM}, who attempted to adapt popular consensus methods based on large crowds to work with smaller sets of experts, discussed the impact on expertise labelling as a result of SMEs having different levels of expertise, utilities, knowledge and subjective preference. These various aspects can prove troublesome for those administering the task, as it is often unknown who is right when there are disagreements. Those with specialist knowledge also often use every piece of information available to them, and this cannot always be accounted for within a machine learning labelling task, which is often a replication of a real-world setting. Lacking this information, experts may then have to pivot making probable guesses, which results in more noise in the data set\cite{Brodley1999IdentifyingMT}.
\par
Whereas some of the previous work has focused on independent aspects of the labelling task like: the number of workers, the number of labels, sample size and the expertise levels of the workers, the differing permutations of each of these parameters has not been fully considered before. More importantly, whether certain ground-truth inference methods are better with certain parameters of a labelling task is not well understood. Previous work has suggested that while there are a number of different options for practitioners to choose from in terms of methods and algorithms, they can largely vary in their performance depending on the data set they are attempting to infer ground-truth from\cite{Zheng2017TruthII}. For the number of workers, previous studies have attempted to offer optimisation for recruitment\cite{Carvalho2015HowMC}, but these have not controlled for the other parameters of the labelling task and have not considered if some methods could provide an opportunity to make the crowd-sourcing recruitment even more efficient. In terms of number of labels, binary label tasks have often been a popular choice in research\cite{Valizadegan2013LearningCM,Zhang2011LearningFI,Welinder2010TheMW}, and while on rare occasions there are tasks with more labels\cite{Yan2010ModelingAE}, these often do not look at data sets where there is a double-digit label taxonomy.
\par
It is also important to understand the scaling of ground-truth inference depending on the number of labels in a taxonomy, as tasks theoretically become more inherently difficult when more label choices are added. Sample size is something that is often overlooked in machine learning research\cite{Balki2019SampleSizeDM}, and time limitations and cost can often prevent large amounts of data being gathered, meaning there is no guidance for practitioners to maximise their chances of getting the most accurate label set with respect to these restrictions. Expertise can be difficult to ascertain, but is valuable knowledge for any practitioner, as the ideal situation is to get the best possible workers to give more confidence in their answers. However, when this is not an option, as can be seen in crowd-sourcing tasks, it can be of benefit to know how many workers should be employed in order to mitigate against this risk. Alternatively, it is also beneficial to know if there are any advantages to ensuring expertise of workers is high before administering a task.
\par
The above summarises some of the important considerations regarding interactions between the parameters in labelling tasks. This paper investigates popular methods for inferring ground-truth from multiple workers and aims to explore how they are impacted by the various parameters of a labelling task, drawing conclusions about which methods are suited to which parameters and providing guidance for when they can be proven to be advantageous when label consensus is sought.
Although previous research has looked at comparison of some popular methods, there are no studies, to our knowledge, that consider the impact of all the parameters discussed above or the relationship of ground-truth inference methods with all these parameters. Thus, this paper will attempt to determine the parameters that popular methods for ground-truth inference work best under and will help provide future researchers with guidance as to when the chosen popular methods are appropriate and when they are not. As well as this, while previous studies have focused on the number of workers crowd-sourcing works best at\cite{Carvalho2015HowMC}, we hope to provide more depth to optimal crowd-sourcing choices by studying the relationships between all the parameters with respect to the methods, as opposed to the impact of a single parameter.
\par

\section{Establishing Ground Truth}
\label{sec:headings}

\subsection{Majority Vote}
The easiest and most common way to get the ground-truth label when multiple workers have provided an answer is to use Majority Vote (MV). In its briefest definition, the answers of workers for each input item are treated as votes, with the popular vote being considered the ground-truth.
\par
Take an example where three workers have labelled a single data item, where the label can be either "red" or "blue". The first two workers choose "red", while the final worker chooses "blue". When a practitioner comes to assigning a label to this data item for training a model, the label "red" would be chosen, since more workers have chosen this than “blue”. This means that any mistakes made by a single worker are mitigated. With a binary label choice and an odd number of workers, there is always a guarantee of a consensus label, but outside of these resources and task choices things can become more difficult. Take the previous example and add a third label - "yellow”. Presume each of the three workers has picked a different label; there is now no majority to pick from for this data item. Alternatively, having an even number of workers for the task with a binary option means that there is a possibility of an even 50/50 split in answers. Let us consider a situation where we have four workers as well as the three labels previously mentioned: two of our workers select "red", one worker selects "blue" and the final worker selects "yellow". This leaves the practitioner in a dilemma; the label "red" now has some agreement among the workers, but it is not a majority exceeding 50\%. When no majority is found, the practitioner is left with three potential options:
\begin{enumerate}
    \item Return to the workers to ask if any are willing to reconsider their answers in the hope a majority can be found. The practitioner must be careful to remain objective in this instance so as not to lead the worker into forcefully changing their label, and there is no guarantee that a majority will be reached for all conflicts.
    \item Drop the data point so it is not removed from the final data set. In contrast to point one, this means that no time is wasted in trying to "fix" labels where no majority can be reached, and instead this time can be dedicated to getting more labels. The obvious downside of this approach is that data loss is inevitable, and tasks of higher difficulty- which are more likely to see disagreement - may not be understood by the model.
    \item From one of the selected answers, pick one at random to use as ground-truth. Answer choice can also be weighted depending on how many workers selected it. This means no data is lost, although there is now the potential of added noise in the labels due to the random selection.
\end{enumerate}
\par
One of MV’s key weaknesses is that the expertise of all workers is considered equal across the board, which is a flawed assumption for most real-world tasks. If you consider a situation where you have one highly accurate worker and two highly inaccurate workers who are consistent with each other (with this information not apparent to the practitioner), we would accept the answers of the two inaccurate workers more often than not, compromising our data set. Even in the case of SMEs, expertise levels can vary, and the decisions made by workers can differ when a task has some level of subjectivity. Despite the issues with MV, it is a quick method and does have theoretical grounding in "wisdom of the crowd\cite{10.5555/1095645}. This suggests that, when putting a question to a crowd, the answer will average out to an approximate of the ground-truth. When employing a high number of crowd-sourcing workers, the hope is that enough of them will give the correct answer so that the risk of those who do make an error is reduced. However, the number of people in the crowd has no rule of thumb and having a high number of workers is often not always possible due to expertise requirements or costs.\par

\subsection{CrowdTruth}
CrowdTruth (CT) 2.0\cite{Dumitrache2018CrowdTruth2Q} is an open-source framework designed to sit on top of crowd-sourcing platforms such as MTurk and CrowdFlower, offering an automated solution to inferring ground-truth data\footnote{ https://github.com/CrowdTruth/CrowdTruth-core}. In contrast to MV that enforces agreement between workers, the framework captures inter-annotator disagreement to help resolve ambiguity in the data when disagreements occur among workers. CT considers that there are three main aspects that should be considered when inferring ground-truth the workers, input data (media units) and annotations\cite{Inel2014CrowdTruthMC}. Upon collection of the crowd-sourced labels, metrics are calculated for each of the three components, which in turn have an impact on each other. For example, if there are 20 workers for a single input data unit and 19 of the workers choose label A, while 1 worker selects label B, the worker who chose label B would be penalised heavily for this in their worker quality score, as the high level of agreement for this input data would assign a high quality metric score. Alternatively, if there are ten labels and worker annotation is evenly split among the workers for an input unit, the worker quality penalty would be lower, while the input data quality score for that item would be low. CT is currently in its 2nd iteration, with the improvements over the first version\cite{Aroyo2015TruthIA} centred around the metrics so that low quality workers disagreeing does not indicate the input data is ambiguous, and vice-versa where a low quality input does not indicate a poor worker. A naive example of this could be in a task where dog breeds are labelled and all the input data is blurred, making it difficult for the workers to select a correct annotation. If workers give different annotations, it is an indicator that the input data itself is the issue and the annotations are not a reflection of the worker quality, thus offering a statistic that allows the task administrator to remove problematic input units.\par
Whereas CT provides a number of useful statistics that help give a good overview of various aspects of a task, we are mostly interested in the “Media Unit - Annotation Score” as shown in equation \ref{eq:eq1}, which gives a score for each input sample to determine the confidence that the annotation (label) is associated with that item. It is weighted by a worker quality score algorithm, thus boosting the answers of better workers while penalising the score of those who are considered to have a lower quality, with quality determined from worker agreement across all annotated units by said worker. Previous experiments investigated thresholds of the \(U\) score, which allows the user to make a judgement about which units they want to use for training\cite{Dumitrache2018EmpiricalMF}. However, there is no rule of thumb to choosing a “best” annotation threshold score, and it seems to depend on the data set - with recommendations to experiment with thresholds.\par
\begin{equation}
U\left(s,g\right)=\frac{\sum_{w\in W}{n_{sg}^{\left(w\right)}Q\left(w\right)}}{\sum_{w\in W} Q\left(w\right)}
\label{eq:eq1}
\end{equation}
\(n_{sg}^{\left(w\right)}\) refers to worker \(w\)’s labelling for input unit \(s\), with each input unit having a choice of \(G\) label. This is weighted by the “worker quality score” (\(Q\)) for each worker \(w\) that has supplied an answer for that input unit \(s\). The outcome of this is a ratio of how many workers picked each annotation weighted by their perceived ability - effectively acting like a weighted MV. \(Q\) is the product of two other equations: the worker-media unit agreement - which is the average cosine distance between a single worker’s annotations and the annotations of the input units they have completed - and the worker-worker agreement, which indicates how similar a worker is to all others by calculating pairwise agreement. Both algorithms are weighted by the unit quality score algorithm, which means workers are not penalised heavily if the input item itself seems ambiguous. Full explanations of all the metrics can be found in the work by Dumitrache et al.\cite{Dumitrache2018CrowdTruth2Q}.
\subsection{Expectation Maximisation}
A popular alternative method for inferring ground-truth is Expectation Maximisation (EM), as demonstrated by Dawid \& Skene\cite{Dawid1979MaximumLE}. The method takes initial guesses of the ground-truth labels, usually done with MV, from the answers of each worker \(w\) and then infers the error rate of each worker by determining how many times a worker w selected label \(g\) when \(j\) is the correct one, with \(\ g,\ j\in G\ \) labels within the taxonomy of label choices \(G\). Doing this produces a square matrix for each worker, \(w\), with each row summing to 1. Therefore, the diagonal of each matrix is the rate at how accurate a worker is for each label and forms the basis of EM, in that it determines how consistent a worker is in applying labels. This error rate, along with the marginal probabilities of all labels \(G\), are used to recalculate the ground-truth. It continually loops between two stages - an expectation step (E-step) and a maximisation step (M-step) - until it converges to produce ground-truth results for all labels. In general, the E-step relies upon the equations of the M-step, thus the M-step is often computed first with an initialised guess for parameters. The following discusses the algorithms in regard to this analysis.
\par
\subsubsection{M-step}
Assuming the ground-truth is unknown for the labels, the error rate, seen in equation \ref{eq:eq2}, for each worker and the marginal probabilities, seen in equation \ref{eq:eq3}, for each label \(j\) are computed in the M-step. Within these equations, \(n_{sg}^{\left(w\right)}\) refers to the counts of given answers for each item in the sample from a worker, similar to that used in equation \ref{eq:eq1}. Equation. \ref{eq:eq2} takes the inferred ground-truth labels, \(T\), and the counts of each answer given, \(n\). For the first instance, \(T\) is randomly initialised if not known, usually with a majority vote. Equation \ref{eq:eq3} takes the inferred ground-truth labels \(T\) and divides it by each item in the sample, \(S\).
\par
EM is capable of handling a worker answering the same sample more than once, but if based on an assumption of each worker only answering each item in the sample once, each vector for a worker’s answer to a sample can be considered as one-hot encoded vector of length \(G\). For example, if worker 1 has three label options and chooses the second label, the worker answer vector for that sample is \([0,1,0]\). For error rate estimation, the estimated ground-truth labels are multiplied by the worker answers for each item in the sample, and these are then normalised to produce a square matrix for each worker \(w\), which gives the error rate of each label.
\par
\begin{equation}
{\hat{\pi}}_{jg}^{\left(w\right)}=\frac{\sum_{s}{T_{sj}n_{sg}^{\left(w\right)}}}{\sum_{g}\sum_{s}{T_{sj}n_{sg}^{\left(w\right)}}}
\label{eq:eq2}
\end{equation}
\begin{equation}
\widehat{P_j}=\frac{\sum_{s} T_{sj}}{S}
\label{eq:eq3}
\end{equation}
\subsubsection{E-step}
Finally, the equations in the M-step are used for the E-step, where new probabilistic estimates of the ground-truth labels \(T\) are made, as can be seen in equation \ref{eq:eq4}. This gives a probability of a label being the inferred ground-truth for each item in the sample. EM then loops between these two steps until it converges. Here, \(q\) is introduced to denote the presumed true label in the set for a sample item, while \(D\) refers to the input data of worker answers.
\par
\begin{equation}
p\left(T_{sj}=1\middle|\ D\right)=\frac{\prod_{w=1}^{W}\prod_{g=1}^{G}\left(\pi_{jg}^{\left(w\right)}\right)^{n_{sg}^{\left(w\right)}}P_j}{\sum_{q=1}^{G}\prod_{w=1}^{W}\prod_{g=1}^{G}\left(\pi_{qg}^{\left(w\right)}\right)^{n_{sg}^{\left(w\right)}}P_q}
\label{eq:eq4}
\end{equation}
The original Dawid \& Skene\cite{Dawid1979MaximumLE} method simply infers the ground-truth of the given data. Many modern methods are based upon this method - including most notably Raykar et al.\cite{Raykar2010LearningFC} - which extends the method to produce a classifier. The previously mentioned work of Valizadegan, Nguyen \& Hauskrecht\cite{Valizadegan2013LearningCM} is also noteworthy, as it adapted the work performed by Raykar et al.\cite{Raykar2010LearningFC} to situations where there are fewer experts, focused solely on high expertise tasks where worker resources may be scarce. The work performed by Raykar et al.\cite{Raykar2010LearningFC} is better geared towards a much higher number of workers, as outlined by the authors. Even the resulting error rate for each worker calculated by EM is useful in itself and can be used to determine the cost of each worker, something particularly valuable in crowd-sourcing\cite{Ipeirotis2010QualityMO}.
\par
\section{Methods}
\subsection{Data sets}
For the data, two different sets were chosen, with the distributions of labels used to create data sets for the various tests. With the based data set being \(D\), they are used to create samples sizes for experiments, denoted as \(S=\{50,125,250,500, 1000,2000\}\). Each item in \(S\) can be thought of as the target sample size that needs to be generated for experiments in order to test the impact of \(S\) on the varying consensus methods. Depending on the label distributions, sample sizes can sometimes be slightly higher or lower than the target \(S\) to help fit to the respective distribution.
\par
\subsubsection{Wisconsin Breast Cancer}
The first data set is the Wisconsin Breast Cancer data set\footnote{\protect\url{ https://scikit-learn.org/stable/modules/generated/sklearn.datasets.load_breast_cancer.html}}, which includes 569 samples where the label is either malignant or benign. The data set was selected due to it being readily available and something that is common for binary classification in research. In this instance, we treat malignant as false and benign as true. The data associated with each label contains 39 different features, all of which are continuous, although it should be noted that the input data to each of the algorithms is the label choices of the workers, not these features.
\par
\subsubsection{20 Newsgroups}
The second data set is the 20 Newsgroups data set\footnote{\protect\url{https://scikit-learn.org/stable/modules/generated/sklearn.datasets.fetch_20newsgroups.html}}, which includes 18846 samples in total and 20 different labels. This data set is used here to generate the different label sets by taking subsets of the data based on the number of labels that are required for the experiment. The data set is one of few to be readily available containing so many label choices, while the distribution of labels is fairly balanced for each subset of selected labels and is popular in natural language processing applications of machine learning. The data associated with each label is in text format and this is again not used as part of any of the ground-truth methods.
\par
\subsection{High and low expertise workers}
Each experiment takes a number from set \(W=\ \{3,5,8,10,13,15,18,20,30,40\}\), with each \(w\ \in\ W\) denoting the number of workers that are generated in a particular experiment. All generated workers are assigned an expertise score \(\lambda\) between two boundaries, \(lb<\lambda\ <ub\), where lb \& ub are the lower and upper bounds of the expertise, respectively. The final expertise score set for all desired \(w\) workers is denoted as \(\mathrm{\Lambda}\ =\ \{\lambda^1,\ \lambda^2,\ldots,\lambda^w\}\). For all higher expertise experiments, \(ub\ =\ 0.99\), while \(lb\ =\ 0.51\). These values are set to ensure that every generated worker is getting at least half of the answers correct, regardless of the number of labels in the set to provide a measure of what we consider a high expertise set, with a limit that suggests they will at least have some minor error. For lower expertise experiments, the \(ub\ =\ 0.8\), while the lower bound is found through a lower-bound model finder, which is discussed in-depth below. 0.8 was chosen due to a similar expert generation upper-boundary selected in Ipeirotis et al.\cite{Ipeirotis2010QualityMO}. This is done to produce more noise in the experiment and represent a lower expertise set. Each worker is assigned a randomly selected expertise \(\lambda\) that is between the two boundaries. \(\lambda\) signifies how many answers we want to keep as correct for an individual worker. As an example, if \(\lambda = 0.6\) - and our ground-truth labels are \(T\) - we want 60\% of a worker’s answers to equal their equivalent value in \(T\), while changing 40\% of the answers to not equal their equivalent in \(T\).
\subsection{Lower-bound model finder}
For the lower-bound model finder, 100 different random forest models are trained with the entire base data set \(D^g\subseteq D\) for all label sets, denoted as \(G=\{2,\ 3,\ 5,\ 7,\ 10,\ 13,\ 15,\ 20\}\), which are associated with L random labels all chosen with equal probability of being selected from the set \(g\ =\ \{1,\ldots,G\}\). While a random forest model has been chosen here, any classification model will suffice, and its selection is simply due to it being a popular method that is widely used. The main goal is to produce lb for expertise that can simulate a worker who is, at worst, performing slightly better than random in their label selection. The experiments presume that workers have good intentions to perform well and are not maliciously selecting the incorrect answer, which is why setting lb to 0 does not make sense. Please note, \(G=2\) is the Wisconsin Breast Cancer data set, while \(G\neq2\) are all subsets of the 20 Newsgroups data set.
\par
Based on the selection of \(G\), \(T\) denotes the ground-truth of \(D^g\). For the Wisconsin Breast Cancer data set, no pre-processing was applied to the data, while for the 20 Newsgroups data, a TF-IDF vector representation for each record was used. As an example, for the experiment with the first 4 labels from 20 Newsgroups, all the data labelled \(g\in{1,2,3,4}\) was gathered (therefore all items not labelled \(\{1,2,3,4\}\) are not used) and vectorised. \(L\) random labels are created which match the size of \(T\) with each item being selected from the selection of \(G\). Data is split - 70\% for training and 30\% for testing. Training uses the 70\% split of \(D\) and its associated \(L\), while the remaining 30\% of \(D\) is used to predict new answers \(L\prime\), with \(L\prime\) then compared against the corresponding 30\% of \(T\) to calculate the weighted f1-score. The mean of all weighted f1-scores for all 100 repetitions is then computed and used as the lb expertise for the experiment. For each \(G\) label set experiment, this lower-bound model is computed once and then used across all permutations where there is the selection of \(G\) labels. The parameters of the random forest models are the defaults in scikit-learn package (version 0.21.3) on Python\footnote{\protect\url{https://scikit-learn.org/stable/index.html}}.The lb expertise scores for each label set in \(G\) are displayed in the Table \ref{tab:table1}.
\par
\begin{table}
 \caption{Lower-bound expertise \(lb\) for each label set, determined by the average f1-score of 100 repetitions of Random Forest Models on the base data.}
  \centering
  \begin{tabular}{ll}
    \toprule
    \cmidrule(r){1-2}
    G (labels in set)     & Lower-bound expertise – lb \\
    \midrule
    2 & 0.483 \\
    3 & 0.327 \\
    5 & 0.189 \\
    7 & 0.139 \\
    10 & 0.094 \\
    15 & 0.063 \\
    20 & 0.049 \\
    \bottomrule
  \end{tabular}
  \label{tab:table1}
\end{table}

\subsection{Label distributions}
Each generated sample size \(S\) is built from the distributions of labels from \(D^g\). For the 20 Newsgroups data, each \(D\) only includes the labels up to the point of \(G\) and chooses them sequentially. So, for example, when \(G\ =\ 3\), \(D\) is reduced to include only the data that is labelled from the set where each individual label \(g\in\ \{1,\ 2,\ 3\}\), and the distributions of this set are used to generate \(S\) and \(L\). The distributions can be seen in Table \ref{tab:table2}. Please note that, due to the samples and rounding, the result of \(S\) is not always exact. For example, in some instances it may include 251 labels instead of 250 so the distribution of the original \(D\) can be best matched.
\par

\begin{table}
 \caption{The distribution of label choices for each \(D^g\), which are used with \(S\) to generate the \(T\) ground-truth labels.}
  \centering
  \begin{tabular}{l|llllllll}
    \toprule
    & \multicolumn{7}{c}{G (labels in set)}\\
    Label & 2 & 3 & 5 & 7 & 10 & 15 & 20 \\
    \midrule
    0 & 0.373 & 0.289 & 0.17 & 0.12 & 0.083 & 0.055 & 0.042 \\
    1 & 0.627 & 0.353 & 0.207 & 0.146 & 0.101 & 0.067 & 0.052 \\
    2 &  & 0.358 & 0.21 & 0.148 & 0.102 & 0.068 & 0.052 \\
    3 &  &  & 0.209 & 0.147 & 0.102 & 0.067 & 0.052 \\
    4 &  &  & 0.204 & 0.145 & 0.1 & 0.066 & 0.051 \\
    5 &  &  & & 0.148 & 0.102 & 0.068 & 0.052 \\
    6 &  &  & & 0.146 & 0.101 & 0.067 & 0.051 \\
    7 &  &  & & & 0.103 & 0.068 & 0.053 \\
    8 &  &  & & & 0.103 & 0.068 & 0.053 \\
    9 &  &  & & & 0.103 & 0.068 & 0.053 \\
    10 &  &  & & &  & 0.068 & 0.053 \\
    11 &  &  & & &  & 0.068 & 0.053 \\
    12 &  &  & & &  & 0.067 & 0.052 \\
    13 &  &  & & &  & 0.068 & 0.053 \\
    14 &  &  & & &  & 0.067 & 0.052 \\
    15 &  &  & & &  &  & 0.053 \\
    16 &  &  & & &  &  & 0.048 \\
    17 &  &  & & &  &  & 0.05 \\
    18 &  &  & & &  &  & 0.041 \\
    19 &  &  & & &  &  & 0.033 \\
    \bottomrule
  \end{tabular}
  \label{tab:table2}
\end{table}

\subsection{Majority Vote vs. CrowdTruth vs. Expectation Maximisation experiments}
For each experiment a label set \(G\) and a sample \(S\) is chosen, with lb and ub to determine \(\mathrm{\Lambda}\) based on the criteria of whether it is a lower or higher expertise experiment. \(S\) determines the number of \(T\) labels we want to produce based on distributions of \(D\). \(W\) worker sets are tested on each of these experiments, with ten repetitions for each \(W\). Each item \(W\) signifies the number of workers we want to generate answers for. For example, \(W\ =\ 10\) means we generate ten workers with different answers governed by their expertise score \(\lambda^w\), chosen between the boundaries of the experiment. Ten repetitions were chosen to provide a balance between obtaining a good range of scores while not being computationally expensive. Initial experiments were run at higher worker sets, but these did not add any value considering their added computational cost.
\par
For each of the ten repetitions, a different accuracy score and worker generated answers were created, producing a unique noisy version of \(T\) for each worker, denoted by \(A_{sw}\). The noise in the labels is determined by the previously mentioned \(\lambda\) parameter, selected at random for each worker between the two expertise boundaries, with items selected to be changed to incorrect also picked at random. For which label to assign to an item selected to be incorrect, each item in \(G\) is assigned an equal probability of 1/\(G\), and a label is picked at random until \(A_{sw}\neq T_s\). With \(A_{sw}\) computed in an iteration, the matrix is used to infer ground-truth \({T\prime}_m\), via each of the algorithms \(M\): MV, CT and EM\footnote{EM implementation taken from: \protect\url{ https://github.com/dallascard/dawid_skene}} . All algorithms are therefore presented with the same generated worker answers in each repetition. The weighted f1-score comparing \({T\prime}_m\) to \(T\) is then computed for each \(M\). An ANOVA test compared each of the f1-score distribution, created by the ten repetitions, with p values recorded and tested for significance of \(p<0.05\) and \(p<0.005\).
\par
As an example of a single experiment, we consider the lower expertise boundaries for workers, set \(S=500\) and \(G=5\), generating a set of \(T_s\) labels based on the distribution of labels from \(D_g\). Each \(w \in W\) is then looped through for ten repetitions, which can be thought of as creating 3 workers with a different expertise 10 times, then creating 5 workers with a different expertise 10 times, and so on. In a single repetition, the workers are created by assigning them an expertise score \(lb\ >\lambda>0.8\). From the ground truth labels \(T\), a noisy version for each worker is created based on a worker’s assigned \(\lambda\). The concatenated version of all worker’s answers, \(A_{sw}\), is then put through each of the three methods, taking the label with the highest value for each item in \(S\) to produce an inferred ground-truth of \({T\prime}_m\). The results for each method, \({T\prime}_m\), are then compared to T and a weighted f1-score is calculated and recorded. This process is completed nine more times with all f1-scores recorded for each iteration, followed by moving onto the next worker set item in \(W\).
\par
\section{Results}
\subsection{Higher expertise}
\begin{figure}[ht]
    \begin{subfigure}{0.45\textwidth}
        \includegraphics[width=1\linewidth, height=6cm]{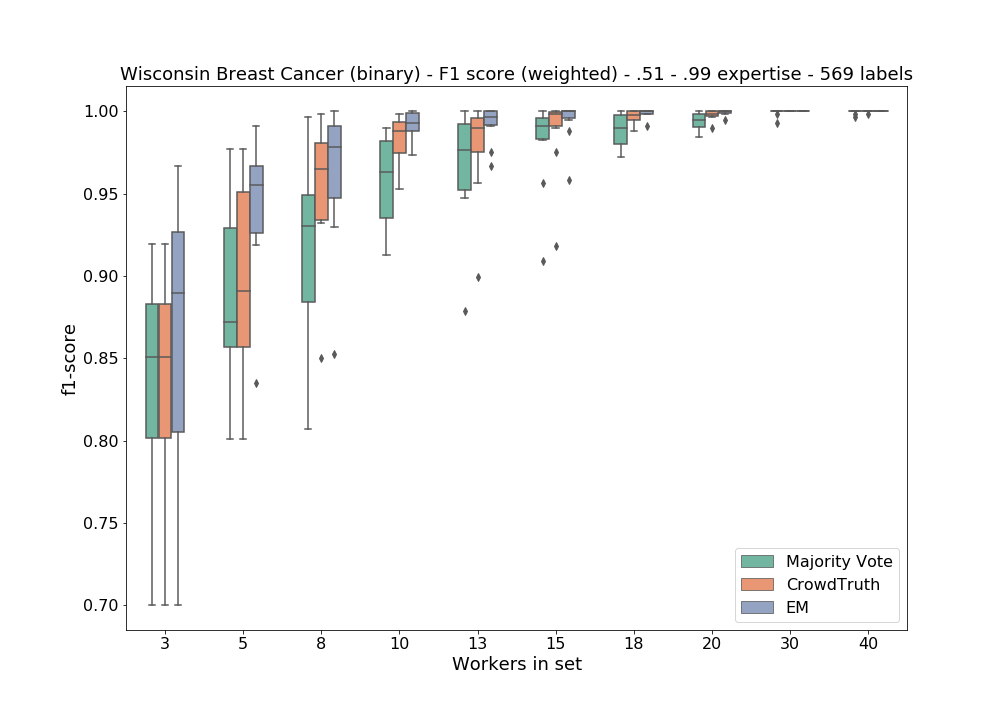} 
        \caption{Higher expertise experiment with two label choices and 569 sample size from the Wisconsin Breast Cancer data set. Note that CT has the same performance as MV when there are three workers.}
        \label{fig:fib1a}
    \end{subfigure}\hspace{0.1\textwidth}
    \begin{subfigure}{0.45\textwidth}
        \includegraphics[width=1\linewidth, height=6cm]{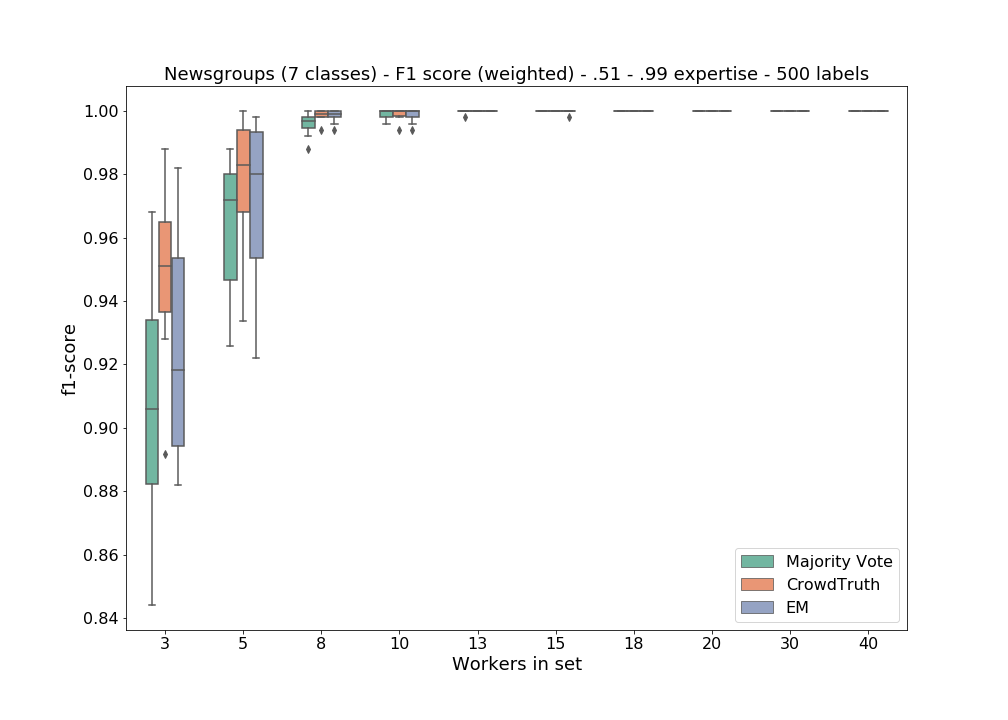}
        \caption{Higher expertise experiment with seven label choices and 500 sample size from the 20 Newsgroups data set. Compared to Fig 1(a), the results converge faster to a near-perfect f1-score with fewer workers.}
        \label{fig:fig1b}
    \end{subfigure}
    \begin{center}
        \begin{subfigure}{0.5\textwidth}
            \includegraphics[width=1\linewidth, height=6cm]{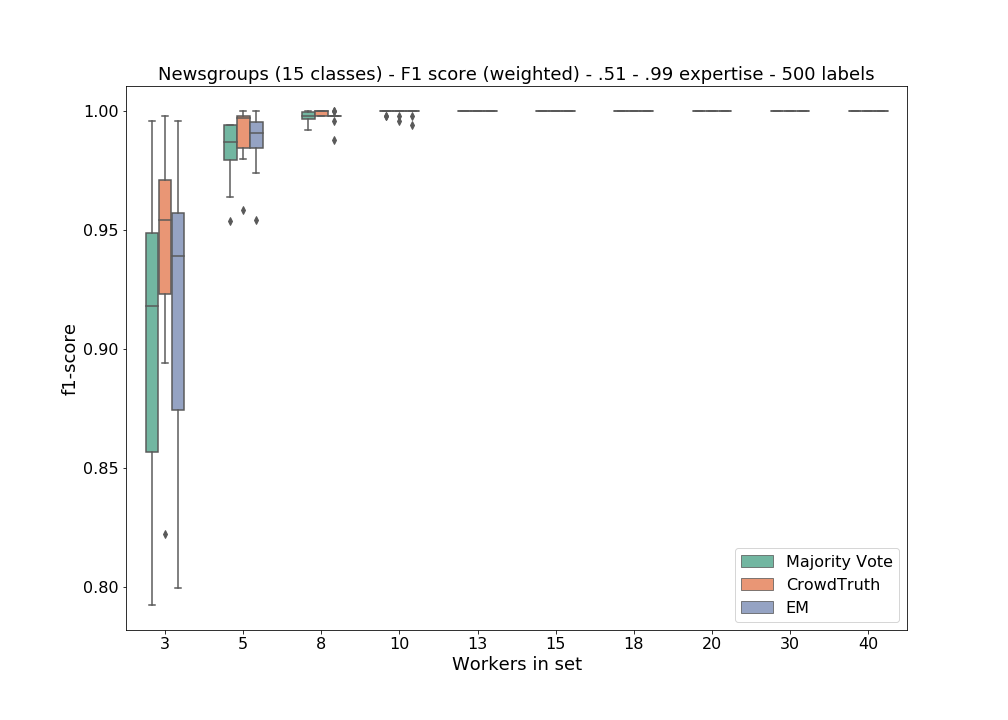}
            \caption{Higher expertise experiment with fifteen label choices and 500 sample size from the 20 Newsgroups data set. While this also converges to a near-perfect f1-score with fewer workers than Fig 1a, it should be noted that no matter how many labels there are to choose from, having three workers will always produce a large variance.}
            \label{fig:fig1c}
        \end{subfigure}
    \end{center}
\caption{Selected results for higher expertise, displaying the quick convergence to near-perfect f1-score for all methods when more labels are included.}
\label{fig:fig1}
\end{figure}

For higher expertise experiments, only\ S\ =\ 500\ experiments. As previously mentioned, all experiments set expertise for each worker \(0.51<\lambda<0.99\). For all experiments, when \(W=3\), all methods had a larger range of f1-scores, with the smallest weighted f1-scores being when \(G=\ 3\ \land W=3\), with these being MV=0.72, CT=0.72 and EM=0.74, although these were all considered outliers, as the mean f1-scores in this experiment instance were MV=0.91, CT=0.95 and EM=0.92. For all experiments using the 20 Newsgroups data, at \(W\geq13\) there is no f1-score <\ 0.99 for all methods, apart from one instance for MV when \(W=13\land G=3\). For experiments where \(\{2,\ 5,\ 7,\ 10\}\subset G\), all results where \(W\geq18\) produced a perfect f1-score of 1.0 for each iteration, while \(\{13,\ 15,\ 20\}\subset G\) also produced perfect f1-scores where \(W\geq13\). Figure \ref{fig:fig1} shows the results of selected experiments, displaying how more label choices resulted in quicker convergence to a perfect f1-score with fewer workers. There were no instances where CT outperformed EM to a significant degree and vice-versa.
\par

\subsubsection{Expectation Maximisation vs. Majority Vote}
Selected results where \(S=500\) are displayed in Fig 2. Green cells below indicate a significant difference of \(p<0.05\) in favour of EM, while red cells indicate instances where there was no significant difference \((p>0.05)\). Of the 70 permutations for the high expertise experiment, EM produced a more favourable result in five instances, with four of them being for \(G=2\), and the remaining being when \(G=3\). Experiment \(G=2\ \land W=10\) produced a significant result where \(p<0.005\).
\par
\begin{figure}[h]
\centering
    \includegraphics[scale=0.3]{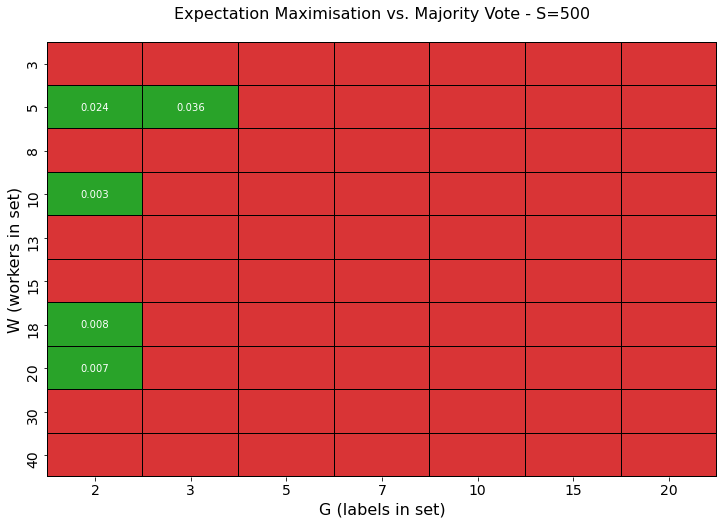}
    \caption{Displaying \(p<0.05\) results where EM outperformed MV for high expertise experiment in green cells. Results where \(p>0.05\) have been omitted in the red cells.}
    \label{fig:fig2}
\end{figure}
\subsubsection{CrowdTruth vs. Majority Vote}
Selected results where \(S=500\) are displayed in Fig 3. Green cells below indicate a significant difference of \(p<0.05\) in favour of CT while red cells indicate instances where there was no significant difference \((p>0.05)\). Of the 70 permutations for the high experiment, CT produced a more favourable result in three instances. For experiments where \(G=2\), experiments where worker sets \(W=10\) and \(W=18\) produced significant results where \(p<0.05\), while \(G=7\land W=3\) produced a single result where \(p<0.05\).
\par
\begin{figure}[h]
\centering
    \includegraphics[scale=0.3]{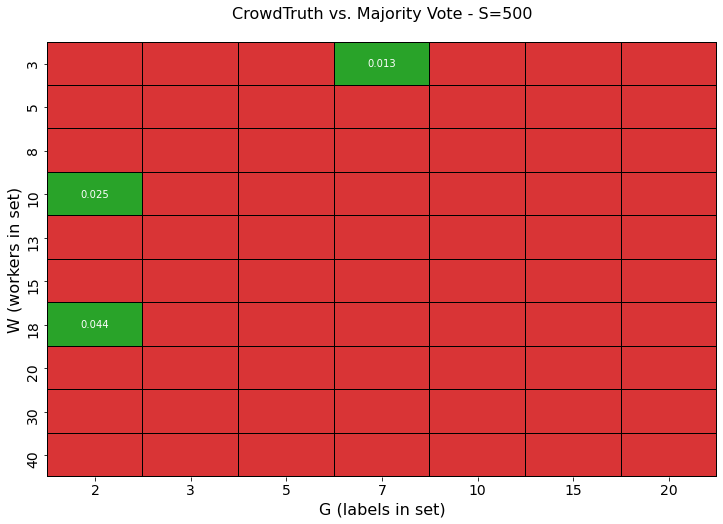}
    \caption{Displaying results where CT outperformed MV with \(p<0.05\) in green cells. Results where \(p>0.05\) have been omitted in the red cells.}
    \label{fig:fig3}
\end{figure}
\subsection{Lower expertise}
For the lower expertise boundaries, where each worker is assigned an expertise of \(lb<\lambda<0.8\), the full set of \(S\) sample size is used, resulting in 420 permutations. Unlike higher expertise, no method converged to perfect f1-scores, although there continued to be a pattern in all experiments where the greater value for W, the higher the average f1-score, while the range of scores for each method tended to get smaller. In all experiments where \(W=3\), CT produced identical results to MV, with every f1-score in the ten repetitions the same.
\par
\subsubsection{Expectation Maximisation vs. Majority Vote}
Among all 420 permutations, EM was more favourable with \(p<0.05\) in 59 (14.0\%) instances. A large proportion (80\%) of these significant results occurred when \(G\le3\), as the likelihood of EM giving a significant advantage over MV falls dramatically as G increases. EM also benefited from a worker set of \(W\geq10\), as only four (6.78\%) of the significant results occurred at the smaller worker sets \((W\le\ 8)\), with the most occurring when \(W=20\), with ten significant improvements (seven of these ten were also when \(G\le3\)). In addition, at \(W=40\), when MV tends to get close to f1-score of 1.0, EM was still able to provide an advantage, although it should be noted again that six of the seven results occurred when \(G\le3\), with the remaining result when \(G=5\). A similar pattern can be seen for S=250, where all seven of the significant results had \(G\le3\). This was 17.5\% of the 40 permutations where \(S=250\). Advantageous performance tended to be more common as \(S\) grew, with both \(S=1000\) and \(S=2000\) sample sizes producing 19 significant results; 47.5\% of all permutations for their respective experiments. Where \(G\geq7\), all three of the significant results were seen when \(S=2000\), with one of these when \(W=18\) and the other two when \(W=20\).\par
Two of the experiments where EM outperformed MV are displayed in Fig 4a and Fig 4b, while one experiment where EM never outperformed MV is displayed in Fig 4c. In Fig 4a, all \(W\geq10\) have a significant improvement of \(p<0.05\) for EM when compared to MV, with \(W\geq20\) also being significant to \(p<0.005\). In Fig 4b, for all \(8\le W\le30\), EM is a significant improvement to a degree of \(p<0.05\) compared to MV. Finally, Fig 4c shows \(G=10\land S=2000\) experiments where EM never outperformed MV, which can be seen by boxplots being a similar size or bigger to their MV counterparts. Overall, there were 17 experiments where \(p<0.005\), with 12 of these occurring when \(W\geq15\land G=2\). Experiments where \(W=40\land G=2\land S\geq250\) produced four significant improvements where \(p<0.005\).\par

\begin{figure}[h]
    \begin{subfigure}{0.45\textwidth}
        \includegraphics[width=1\linewidth, height=6cm]{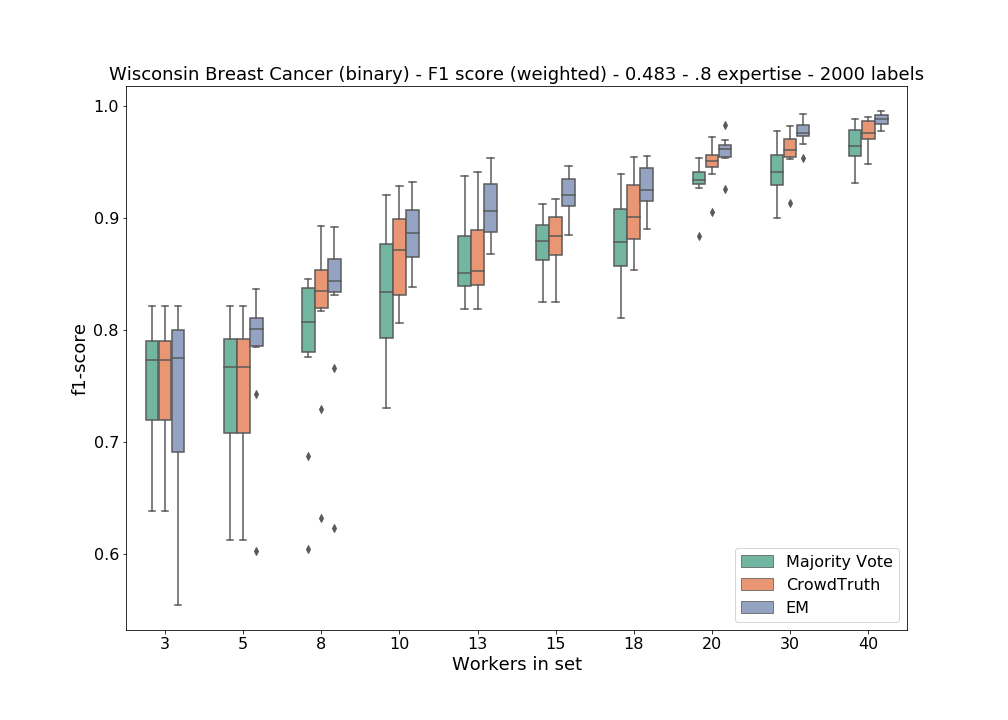} 
        \caption{Lower expertise experiment with two label choices and 2000 sample size from the Wisconsin Breast Cancer data set. The f1-score for EM when there are ten or more workers in a binary data set always outperforms MV.}
        \label{fig:fib4a}
    \end{subfigure}\hspace{0.1\textwidth}
    \begin{subfigure}{0.45\textwidth}
        \includegraphics[width=1\linewidth, height=6cm]{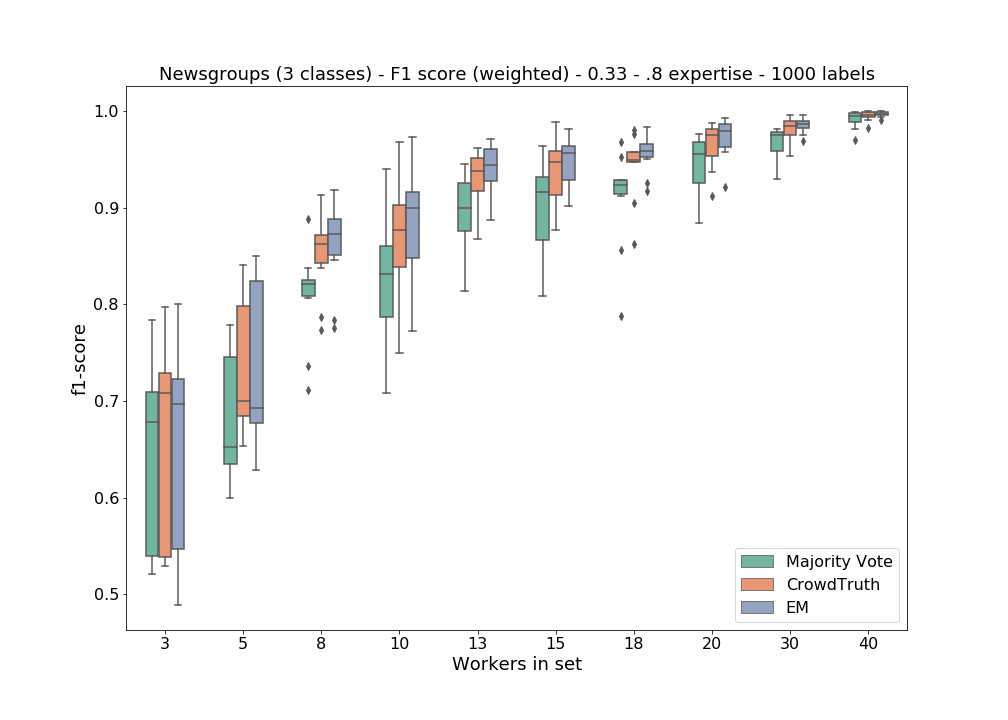}
        \caption{Lower expertise experiment with three label choices and 1000 sample size from the Newsgroups data set. With between 8 and 3 and 30 workers, EM outperforms MV.}
        \label{fig:fig4b}
    \end{subfigure}
    \begin{center}
        \begin{subfigure}{0.5\textwidth}
            \includegraphics[width=1\linewidth, height=6cm]{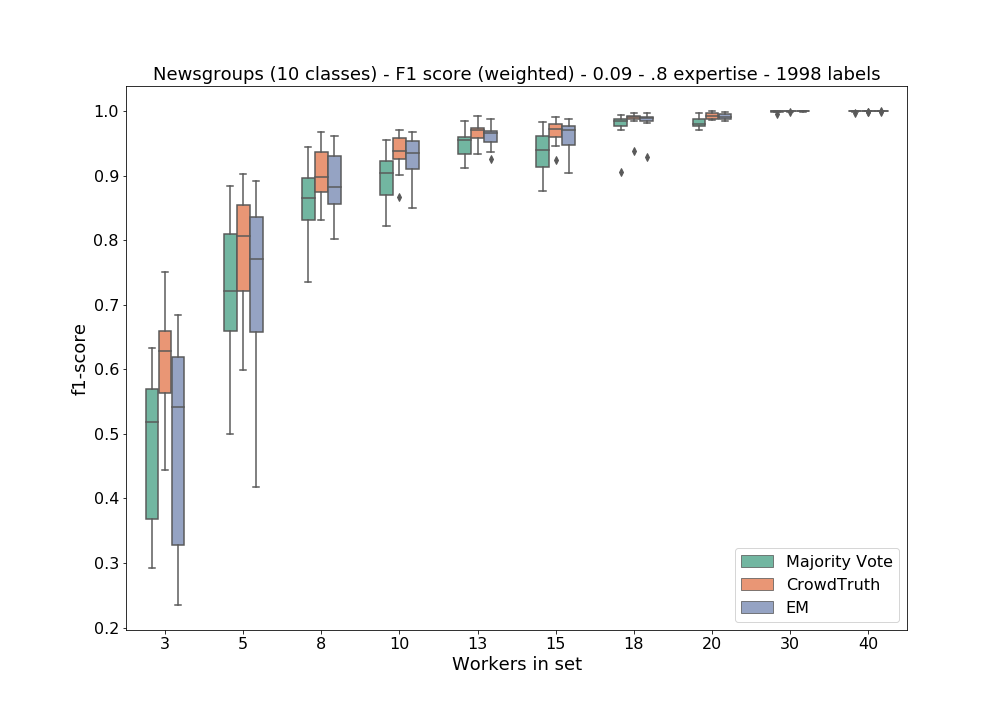}
            \caption{Lower expertise experiment with ten label choices and 1998 sample size (S=2000) from the 20 Newsgroups data set. Here, EM did not manage to outperform MV, which was common for many of the lower expertise experiments that had a large label choice size.}
            \label{fig:fig4c}
        \end{subfigure}
    \end{center}
\caption{Charts displaying 2 instances where EM performed well, and one where it did not offer any real improvement over MV.}
\label{fig:fig4}
\end{figure}

\subsubsection{CrowdTruth vs. Majority Vote}
For 55 (13.1\%) of the 420 permutations CT had a significant advantage over MV, a similar number to EM, although the instances where \(p<0.05\) against MV tended to differ. Of the experiments that returned a significant result all occurred between \(2\le G\geq10\) apart from one which appeared when \(G=20\). The incidence of \(p<0.05\) experiments for \(G\) grows as there are more labels added, before peaking at \(G=5\) when the number of significant results is 14, and this is then followed by a drop until \(G=15\) where there are no results where \(p<0.05\). 58.9\% of all significant results occurred when \(G=5\vee G=7\). Occurrences of \(p<0.05\) experiments tended to rise steadily as S grew where \(50\le S\le500\), until \(S=1000\) and \(S=2000\) indicates a similar number of significant occurrences compared to \(S=500\). Overall, there were six results where \(p<0.005\).\par
The pattern in \(G\) differs from EM which observed a much steeper drop in significant results as \(G\) grew, as can be seen below in Fig 5a. The pattern for significant occurrences in \(S\) and \(W\) is somewhat similar between the two methods when comparing f1-score distributions to that of MV - although for \(S\), EM appears to benefit more from a larger sample size, as can be seen in Fig 5b.\par

\begin{figure}[h]
    \begin{subfigure}{0.45\textwidth}
        \includegraphics[width=0.9\linewidth, height=5cm]{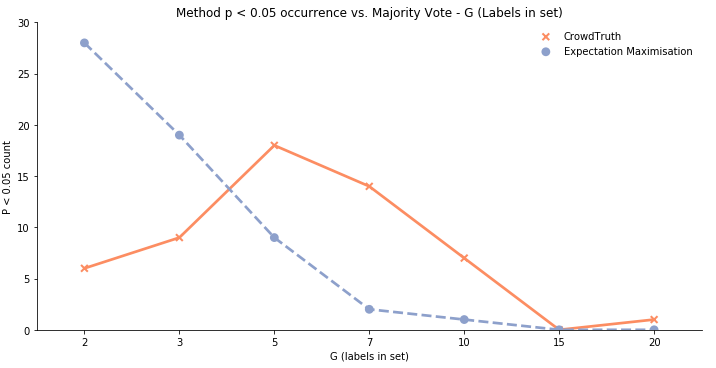} 
        \caption{Showing instances where EM and CT f1-scores outperformed MV with respect to the number of label choices. EM is much better suited to fewer label choices, while CT peaks when there are five label choices. Neither perform that much better with a large label set choice.}
        \label{fig:fib5a}
    \end{subfigure}\hspace{0.1\textwidth}
    \begin{subfigure}{0.45\textwidth}
        \includegraphics[width=0.9\linewidth, height=5cm]{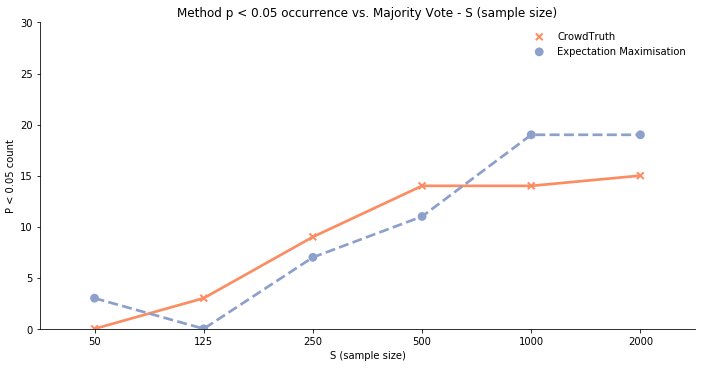}
        \caption{Showing instances where EM and CT f1-scores outperformed MV with respect to the sample size. CT and EM are more likely to produce an advantage over MV as the sample size grows.}
        \label{fig:fig5b}
    \end{subfigure}
    \begin{center}
        \begin{subfigure}{0.5\textwidth}
            \includegraphics[width=0.9\linewidth, height=5cm]{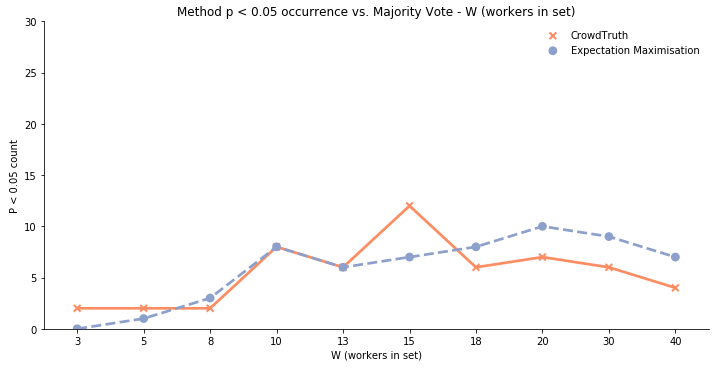}
            \caption{Showing instances where EM and CT f1-scores outperformed MV with respect to the number of workers in the set. CT and EM have a similar pattern. Neither are strong in performance compared to MV when there are few workers.}
            \label{fig:fig5c}
        \end{subfigure}
    \end{center}
\caption{Charts displaying counts of significant experiments for labels in set, sample size and workers in set when comparing results of EM and CT against MV.}
\label{fig:fig5}
\end{figure}

\subsubsection{CrowdTruth vs. Expectation Maximisation}
Overall, 27 (6.4\%) of the 420 experiments produced a result where using CT or EM was more favourable than the other, with 14 of these for CT and 13 for EM. However, of the 14 for CT, only 6 of these were results where CT was also more advantageous than using MV. All 13 of the EM results where \(p<0.05\) against CT were also results where \(p<0.05\) against MV, which also all occurred when \(G=2\ \land W\geq13\). Five of the 14 results where CT was better than EM occurred when \(S=50\). Overall, there were 1.4\% of instances where CT was the standout method to use and 3.1\% of instances where EM was the standout method. In no experiments did MV come out as the clear winning method.\par

\section{Discussion}
Whereas CT and EM do offer advantages over a simple MV, this depends on the parameters of the labelling task. EM is more likely to produce better results if the label set is binary, whereas for CT it is harder to determine a general rule of thumb for when it is the outright best method to use, although it does seem effective with moderately sized label sets. Neither seems to give an advantage for large label sets, while the same is true for small worker sets, where a large variance in f1-score is seen for all methods. The experimental results show the relationship between the various parameters and these particular methods, and give a new understanding as to when they actually provide an advantage.\par
Overall, the results presented here should offer practitioners a general guidance of when to utilise one of these methods rather than MV depending on their choices of workers, expertise, labels and sample. Caution is, however, advised in that this is not a recommendation to base a label task around using a particular method, or just choose one without consideration, but rather seeing where there is an opportunity to use one once a label task has been completed. While previous research has looked at the optimal number of workers for ground-truth inference with multiple workers\cite{Carvalho2015HowMC},  the results suggest that the other task parameters can be just as important when attempting to maximise accuracy through methodology choice. Since label gathering can often be considered the most crucial aspect of any supervised machine learning project, the insight offered here provides practitioners with a huge benefit when attempting to get the best possible data in an efficient manner.\par
If worker ability can be assured to be of high quality, then fewer workers will be required to gain a near perfect ground-truth consensus, while the method choice is not as crucial when compared to lower expertise workers. Considering only three workers, somewhat favourable labels are gained when all can be assured to get at least 51\% of answers correct when compared to the ground-truth. Although results get slightly better the more labels that are added, we often observe convergence to a perfect score with eight to ten workers, with anything beyond this not adding any real value. In the situation that worker cost is high, our analysis indicates there is an advantage in pre-screening workers and gauging their ability, as a small quality check can remove noisy workers and focus on a team with more confidence in their ability. However, the cost of this screening and the hit rate of getting good workers needs to be taken into consideration, while any task to perform this needs to be balanced enough that the expertise scored garnered for workers has a high degree of confidence. One interesting approach from any screening experiment would be to group workers based on the screen into two separate groups: “strong” and “weak”, to see performance on a full task and record the impact on the full task results.\par
With regards to lower expertise workers, we see that smaller numbers result in more variance for the final f1-score for all methods, as less than eight workers often means that the final results can become quite unstable. In addition, a high number of workers often results in convergence to a near-perfect f1-score for all methods, although even in these instances there are occasions with lower expertise workers where CT and EM can still provide an advantage over MV. Variance in the result does tend to reduce as the number of workers grow, although this is not as sudden for the lower expertise worker sets as can be seen when compared to the higher expertise experiments.\par
Both methods tended to produce their best results when there were ten or more workers, although incidents of better performance did not tend to grow in parallel with the number of workers, which could be attributed to other parameters having a bigger impact, or the fact that more workers, on average, tended to produce more accurate results due to the nature of crowd averages phenomenon. Whereas previous research has suggested that ten or eleven workers is the optimal number regardless of other factors\cite{Carvalho2015HowMC}, the results here demonstrate that employing slightly more workers than this can still offer some gains when the expertise of workers can drop below 50\%. For smaller worker sets, often seen in specialist knowledge settings where there are fewer available resources\cite{Valizadegan2013LearningCM}, the results indicate the challenges that they face, in that the final consensus results can vary to a large degree, and none of the methods used in these experiments offer a robust solution to inferring ground-truth.\par
For binary tasks, or even some tasks with three labels in the taxonomy, the choice of EM is clear, as it can often be seen to provide an advantage when this label parameter is met. If the label count is low, there can be reasonable assurance that EM will offer an advantage in inferring ground-truth, so it is recommended to consider it. The method does not appear to scale as well to the number of labels in a set; this causes an issue for real world applications, where large label taxonomies are common due to the complex problems practitioners attempt to solve, as the potential advantages are sparse outside of tasks with fewer than seven labels.\par
On the other hand, CT is a method that generally offers advantages over MV across a broader range of task parameters - although it can be hard to provide a general rule of thumb of when it is best used, unlike EM. Whereas it does offer some advantages at smaller label taxonomy sizes, most of its performance enhancements over MV are seen with slightly larger taxonomies where there are five to seven labels to choose from. One of the curious aspects of CT is seen in the lower expertise experiments for binary label tasks and three workers, as it produced exactly the same results as MV every time, suggesting that it mimics MV with the minimum number of workers (three) and labels (two) for a task, rendering it redundant. Recommending it as a way of improving results over both methods is also tricky, as there were very few occasions when it was the standout method. When it was, these tended to be sparsely populated across the parameter experiments. It is clear that CT is a useful tool and can offer advantages in inferring ground-truth, but giving a definitive idea of when to use it to benefit from above advantages is not entirely obvious.\par
Overall, we discovered that large label taxonomies tend to mean that the choice of method for ground-truth inference is redundant. This may be due to some relation to the class subset sample size, as it can be seen from the overall sample sizes that CT and EM both have some sensitivity to this. Keeping in mind that it is usually the goal to accrue as much data as possible, EM seems to be even more sensitive to the overall sample size, concluding that 500-1000 as a minimum sample size can increase the possibility that a particular method will help boost ground-truth inference. This is particularly curious as the example in Dawid \& Skene\cite{Dawid1979MaximumLE}  base their work on a, albeit contrived, sample size of 45.\par
Considerations over sample size are particularly true for CT where it is recommended to use thresholds on their media unit annotation score to remove noisy examples, which can help aid training data for a model\cite{Dumitrache2018EmpiricalMF}. Thresholds that remove too many labels could, in theory, be costly to CT, meaning the method is unlikely to offer substantial benefit over a standard MV when compared to the results of experiments conducted here. Further experimentation with even larger sample sizes could prove beneficial in understanding whether the occurrences of significant improvements increases in likelihood with respect to the other parameters, or if it tails off to a similar number for each sample set, as can be seen in sample sets 1000 and 2000 for both CT and EM.\par
This also raises questions about label distributions and minority classes, in particular how both of these methods handle it and if some instances of beneficial performance could actually be explained by sub-samples of each label. As we see in much research, the two data sets used here tended to have much more desirable distributions than many real-life equivalents, and analysis into the impact of label imbalance could add to the knowledge acquired here in informing others on how to infer consensus in this particular scenario. Indeed, it could be theorised that, since both EM and CT produce statistics related to each label, the same statistic between a minority class and a majority class is not entirely fair. Whereas these experiments have focused on overall performance, examining the performance of each method on each class with controls on class distributions where minority classes are introduced is a possible avenue for future research to assess how robust each method is when it is presented with these issues. Taking EM as an example, the error rate square matrix produced for each worker will likely be impacted by imbalanced label distributions, and any gains gathered in the overall picture may be massively skewed by the dominant classes, whereas minority classes are often included due to their high importance.\par

\section{Conclusions}
Here, we have concluded that each method has its advantages and draw backs. However, as a first choice we recommend:\par
\begin{enumerate}
    \item EM is a better choice for smaller label sets, while CT is more favourable in medium-sized label sets. Neither method offers much advantage when the label set size is large.
    \item Employing 10-20 workers will not only result in a good balance between overall accuracy and efficiency, but also increases the likelihood that CT or EM can produce more accurate ground-truth inference than MV.
    \item None of the methods can provide much advantage when there are fewer than ten workers, while accuracy also tends to be less stable, so alternative methods should be sought to mitigate any potential errors in the label sets.
    \item High average worker expertise can result in needing fewer workers to gain a better consensus score, although this means methodology for inferring ground-truth is not as important.
    \item Practitioners should aim to get each of their workers to provide an answer to as much sample as possible if they want to reap the potential advantages of CT or EM, as both benefit from a larger sample.
\end{enumerate}
\hspace{\parindent}From the experiments, a loose rubric on these popular methods is now available to help not only guide data-gathering exercises, but also inform others of opportunities to enhance ground-truth inference when the parameters of their exercise align with those in the experiments of this paper. In this paper we have seen that CT and EM are only likely to be advantageous with certain parameter conditions, which was not previously understood. The results provide researchers with new insight into the parameter relationships and how they might impact accuracy on a task. This is particularly important for any supervised learning task, as accuracy of the labels associated with the data is crucial to the success of any model, and every opportunity to increase this accuracy should always be taken.\par
The experimental set up in this paper can help guide future research in testing other methods to see the parameter conditions they work best under, as well as providing researchers with a framework to establish new methods that can attempt to resolve some of the shortcomings of the methods used in these experiments.\par

\bibliographystyle{unsrt}  
\bibliography{references}

\section*{Supplement}
Results from the experiments in this paper can be found at: \protect\url{https://github.com/Caspian-Ltd}

\end{document}